\documentclass{article} 
\usepackage[final]{neurips_2025} 

\PassOptionsToPackage{table}{xcolor}
\PassOptionsToPackage{hyphens}{url}

\usepackage{booktabs}
\usepackage{longtable}
\usepackage{array}
\usepackage{colortbl}
\usepackage{multicol}
\usepackage{multirow}
\usepackage{threeparttable}
\usepackage{graphicx}
\usepackage{xcolor}

\definecolor{topgreen}{HTML}{D8F0B7}

\newlength{\captionskipamount}  
\newlength{\floatskipamount}    
\usepackage{caption}
\usepackage{subcaption}

\usepackage{listings}
\usepackage{xcolor}

\lstdefinelanguage{json}{
    basicstyle=\ttfamily\small,
    numbers=left,
    numberstyle=\tiny\color{gray},
    stepnumber=1,
    numbersep=6pt,
    showstringspaces=false,
    breaklines=true,
    frame=single,
    backgroundcolor=\color{gray!5},
}

\usepackage[utf8]{inputenc}
\usepackage[T1]{fontenc}

\usepackage[sort,compress]{cite}

\usepackage{hyperref}
\hypersetup{
  colorlinks=true,
  linkcolor=red,
  citecolor=cyan,
  filecolor=magenta,
  urlcolor=blue,
}
\usepackage{url}
\usepackage{amsfonts}
\usepackage{amsmath}
\usepackage{amssymb}
\usepackage{nicefrac}

\renewcommand{\footnoterule}{%
  \kern-3pt%
  \hrule width 0.4\columnwidth height 0.4pt%
  \kern 2.6pt}

\usepackage{microtype}
\usepackage{sourcesanspro}

\usepackage{xspace}
\usepackage{enumitem}
\usepackage{textcomp}
\usepackage{stfloats}
\usepackage{verbatim}
\usepackage{wrapfig}
\usepackage{graphicx}

\usepackage{tikz}
\usepackage{algorithm}
\usepackage{algpseudocode}
\usepackage{pgfplots}
\pgfplotsset{compat=1.18}

\def\ie{\emph{i.e., }}

\usetikzlibrary{positioning}

\usepackage{placeins}
\title{
\texorpdfstring{
  Matrix-Game 4.0: Harnessing Video Models for Real-Time Playable Game Worlds
}{
  Matrix-Game 4.0: Harnessing Video Models for Real-Time Playable Game Worlds
}
}

\author{
Jihai Zhang\textsuperscript{1,2}\thanks{Equal contribution.}\quad
Runjia Qian\textsuperscript{1*}\quad
Kaiqin Kong\textsuperscript{3,1*}\quad
Biao Jiang\textsuperscript{1*}\quad\\
\textbf{
Puyi Wang\textsuperscript{2}\quad
Siru Chen\textsuperscript{4,1}\quad
Tianle Li\textsuperscript{2}\quad
Guanjie Chen\textsuperscript{5,6}\quad
Junyi Wu\textsuperscript{5}\quad
}\\
\textbf{
Chubin Chen\textsuperscript{1}\quad
Yuanhang Li\textsuperscript{1}\quad
Kaichen Huang\textsuperscript{1}\quad
Haobo Zhang\textsuperscript{1}\quad
Yangguang Li\textsuperscript{1}
}\\
\textbf{
Wei Yu\textsuperscript{1}\thanks{
  Project Lead and Corresponding Author.
  \newline Technical Report.
}\quad
Guibin Chen\textsuperscript{1}\quad
Yu Cheng\textsuperscript{7,1}\quad
Yahui Zhou\textsuperscript{1}
}
\\[1em]
\textsuperscript{1}SkyworkAI
\qquad
\textsuperscript{2}The Chinese University of Hong Kong
\qquad
\textsuperscript{3}UC San Diego
\\
\textsuperscript{4}University of Science and Technology of China
\qquad
\textsuperscript{5}Shanghai Jiao Tong University
\\
\textsuperscript{6}Shanghai AI Lab
\qquad
\textsuperscript{7}Nanyang Technological University
\\[1em]
Project page:
\href{https://matrix-game-v4-0.github.io/page/}
{\textcolor{blue}{Matrix-Game-4.0}}
\quad
Code:
\href{https://github.com/SkyworkAI/Matrix-Game}
{\textcolor{blue}{GitHub}}
\quad
Checkpoints:
\href{https://huggingface.co/Skywork/Matrix-Game-4.0}
{\textcolor{blue}{Hugging Face}}
}

\begin{document}
\maketitle

\noindent\begin{minipage}{\linewidth}
  \includegraphics[width=\linewidth]{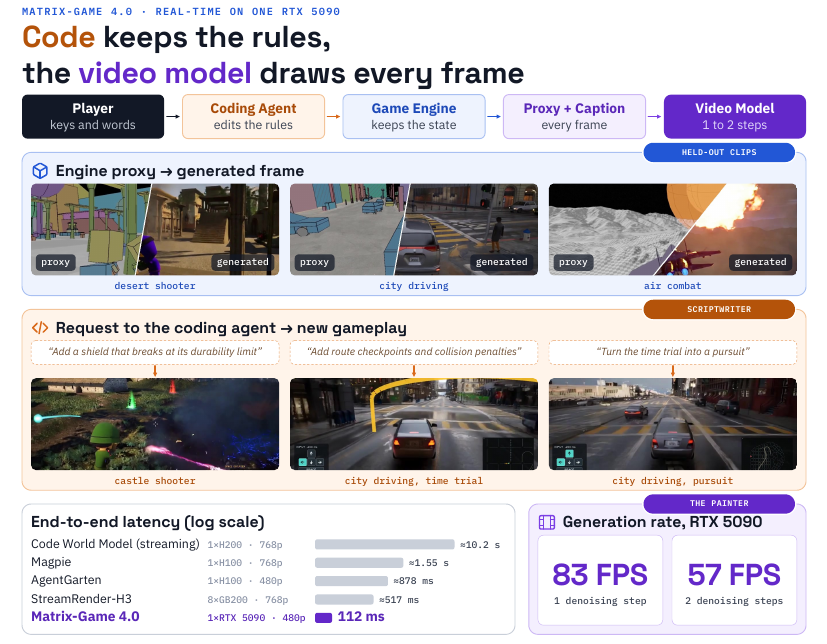}
  \captionof{figure}{\textbf{Matrix-Game 4.0.} 
  From top: the system loop; diagonal split-screen comparisons between low-poly engine outputs and neural renderings on held-out validation examples; gameplay examples from games created by the coding agent in response to user requests; end-to-end latency comparisons with recent systems (Table~\ref{tab:compare}) and the rendering throughput at 480p on a single RTX 5090.}
  \label{fig:teaser}
\end{minipage}
\newpage 

\begin{abstract}
Moving from controllable videos to genuinely playable worlds requires persistent world state, executable game rules, and low-latency interaction. We outline a roadmap for the next generation of Matrix-Game that separates game execution from visual generation through an executable game harness, and introduce Matrix-Game~4.0 as the first system built on this design. A lightweight game engine maintains world state and executes interaction rules, a multimodal coding agent translates player ideas into game code and dynamic scene captions, and a renderer based on autoregressive video model transforms engine-rendered low-poly proxies into detailed visuals, conditioned on dynamic captions and memory references. This framework provides continuous guidance for visual generation and enables natural-language creation and modification of scenes and gameplay. Our renderer is trained on aligned low-poly and high-fidelity video pairs and distilled into a causal generator that generates in four-frame chunks with one or two denoising steps. Together with targeted inference optimizations, Matrix-Game~4.0 enables 480p, 60~FPS gameplay on a single NVIDIA RTX~5090 with a key-press-to-display latency of 112~ms. Single-step generation achieves 83~FPS steady-state throughput with 18.2~GiB of peak GPU memory usage. With fixed prompts, steady-state throughput also reaches 67~FPS on an RTX~5080 and 58~FPS on an RTX~5070~Ti. We demonstrate customizable shooter and driving games with engine-enforced mechanics, and release code, model weights, and low-poly game assets to support research and creation of playable game worlds.
\end{abstract}


\section{Introduction}
\label{sec:intro}

Interactive world models such as the Genie series~\cite{bruce2024genie,parkerholder2024genie2,genie3} extend generative modeling from producing videos for passive viewing to creating worlds for active interaction, demonstrating that interactive environments can be generated on the fly~\cite{valevski2024diffusion,oasis2024,he2025matrix,wang2026matrixgame3,qian2026matrix}. However, as the goal shifts from brief exploration to long-running games with rich gameplay and controllable rules, systems that rely primarily on video generation to maintain world state face four key challenges.

\begin{itemize}
    \item \textbf{No continuous external correction.}
    Model rollouts proceed largely in isolation from external environments. World state exists implicitly in generated pixels and model context, without explicit state being continuously maintained for regions outside the field of view.

    \item \textbf{Accumulating errors.}
    Without external correction, small errors compound through autoregressive generation, progressively degrading consistency. Even with external 3D tools, errors in estimated camera poses can accumulate and eventually cause scene collapse.

    \item \textbf{Uncompressed frame-based memory.}
    Historical frames accumulate without their spatial information being consolidated into a compact representation. Memory costs therefore grow with rollout length, limiting long-horizon generation.
    
    \item \textbf{Limited gameplay mechanics.}
    Interaction centers on navigation and basic environmental responses, with limited support for executable rules, persistent action consequences, and explicit objectives that users can actively pursue.
\end{itemize}

More fundamentally, diffusion models primarily learn statistical correlations: visual realism does not necessarily imply causal or physical understanding. Following Roman Jakobson's tripartite classification~\cite{jakobson2021linguistic}, the language-to-visual mapping in text- and image-to-video (TI2V) generation can be viewed as \emph{intersemiotic translation}. We argue that current video generators are not yet \emph{vision foundation models} for world modeling: much of their semantic understanding and apparent intelligence comes from upstream  multimodal language models (MLLMs) or text encoders, while diffusion acts more like a renderer. Yet addressing these limitations alone is insufficient: genuinely playable worlds also require real-time generation with low interaction latency, meaningful game mechanics and rich interactions beyond navigation, and shared world state for multiplayer interaction. These considerations motivate an architecture where an external system maintains persistent world state and executes interaction rules, while the video model focuses on visual rendering.


\begin{figure*}[t]
  \centering
  \includegraphics[width=\linewidth]{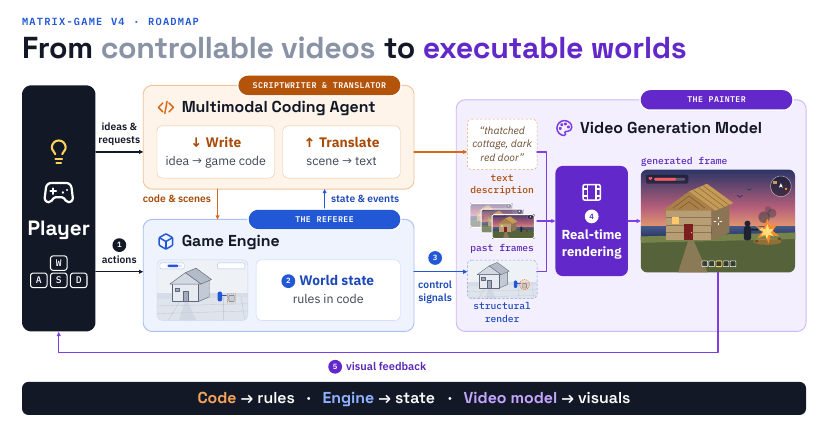}
  \caption{\textbf{Roadmap of Matrix-Game V4}: The multimodal coding agent and the game engine form the harness for the video generation model. The player sends ideas and requests to the agent and actions to the engine (1). The agent writes the world, turning ideas into game code and scenes for the engine, and translates the world, turning the state and events reported by the engine into text descriptions. The engine maintains the world state and the rules in code (2) and sends control signals as a structural rendering of the scene (3). The video generation model renders frames in real time from the text description, its past frames, and the structural rendering (4). The generated frame returns to the player as visual feedback (5).}
  \label{fig:overview}
\end{figure*}

We introduce \textbf{Matrix-Game~4.0}, a framework that places a video generation model within an external \emph{harness} to separate game logic from visual generation. The system has three components (Figure~\ref{fig:overview}). The \emph{multimodal coding agent} translates player ideas into runnable game code and scene descriptions. The \emph{game engine} maintains world state, executes interaction rules, and renders simplified geometry as structural rendering (i.e. low-poly proxy). The \emph{video generation model} transforms this proxy into detailed frames with materials, lighting, and atmosphere. At runtime, player actions update the engine state; the engine outputs proxy frames and structural, motion, and dynamic semantic conditions; and the video generation model renders the detailed frames displayed to the player. These continuous external guidance from the game engine provides a basis for correct rendering, such that the game no longer relies on the video generation model's memory to maintain world state or execute rules. Our main contributions are as follows:
\begin{itemize}
    \item \textbf{Real-time, playable video generation on a single consumer GPU.}
    Matrix-Game~4.0 enables 480p, 60~FPS gameplay on a single NVIDIA RTX~5090, with 112~ms of key-press-to-display latency. Single-step denoising achieves 83~FPS steady-state generation throughput with approximately 18.2~GiB of peak GPU memory usage, making video-model-driven games practical for local deployment and real-time interaction. (Table~\ref{tab:compare}) With fixed prompts, single-step generation also reaches approximately 67~FPS on an RTX~5080 and 58~FPS on an RTX~5070~Ti, extending deployment to consumer GPUs with 16~GiB of memory.

\item \textbf{A roadmap toward playable worlds.}
We outline Matrix-Game V4's roadmap, separating game execution from visual generation through an executable harness. It combines a lightweight engine for world-state maintenance and rule execution, a multimodal agent for creating and modifying gameplay and scenes, and an autoregressive video model for real-time rendering, advancing from controllable videos to playable and extensible worlds.

    \item \textbf{A reusable pipeline spanning data, training, and runtime.}
    We integrate data collection of paired proxy-detailed frames and dynamic captions, model training for video rendering, few-step autoregressive distillation, and efficient real-time inference.

    \item \textbf{An open-source foundation for generative games.}
    We open-source Matrix-Game~4.0 at launch. For researchers, it provides
a reusable technical foundation for further exploration of this
emerging paradigm, combining executable game harnesses with
video-model-based rendering. For game developers, it offers a
concrete preview of how coding agents and neural rendering can
support future game-development workflows, while lowering the
barrier to prototyping and customizing scenes, interactions,
and gameplay through natural-language instructions.
\end{itemize}

\section{Roadmap}
\label{sec:roadmap}

Matrix-Game V4 aims to build a truly playable world. Based on the analysis above, our approach is to wrap the video model in a harness that maintains world state and provides an \textbf{executable foundation for gameplay}. This architecture brings together three modules: a game engine, a multimodal coding agent, and a video generation model. The game engine serves as an external game manager, supplying spatial layout references, defining game rules, managing world state, and updating the semantic context. The multimodal coding agent bridges players' ideas and game code, enabling flexible creation and modification of both gameplay and scenes. The video model then integrates the resulting context and control signals to autoregressively generate short sequences of realistic future frames. Figure~\ref{fig:overview} shows how the three parts work together.

\subsection{Game Engine: The World's ``Referee''}
\label{sec:roadmap_engine}

The engine keeps track of what actually happens in the world and determines the consequences of player actions. Rather than producing fully detailed visuals, it represents roads, buildings, characters, and interactive objects through simplified geometry and scene structures. Game state and rules---including health, inventory, gravity, collisions, and quest progress---are explicitly maintained and enforced in code, rather than inferred from generated images.

As players act, the engine continuously updates the world and outputs structural renderings. These act as dynamic ``scene sketches,'' providing the video model with constraints on spatial layout, object motion, and camera viewpoint. Even when an object leaves the field of view, its state is still maintained by the engine according to the game rules.

\subsection{Multimodal Coding Agent: The World's ``Scriptwriter'' and ``Translator''}
\label{sec:roadmap_agent}

The Multimodal Coding Agent provides the system's intelligence and imagination, while also constructing its game rules. It communicates with the engine in both directions: translating established scenes and events into visual descriptions, and bringing players' ideas or its own creative proposals into the game world.

\paragraph{Downstream: Translating the World.}
The agent converts engine-defined scene structures and interaction outcomes into text descriptions that guide video generation. With a reference image, a simple box can be described as ``an old thatched cottage weathered by sea winds, with a dark red wooden front door.'' When a player shoots an oil barrel, the engine determines whether the shot hits and triggers an explosion; the agent only describes the resulting visuals.

\paragraph{Upstream: Writing and Developing the World.}
The agent adds content and rules to the engine in two ways.
{When players have a specific idea, the agent turns it into a working game.} From reshaping terrain and placing objects to modifying NPC behavior and designing new mechanics, it implements requests by editing scene configurations and game code. For example, ``make monsters swarm in from all directions'' requires more than adding monsters: it also means defining where they spawn, how they move and attack, and what determines victory or defeat.
{When players leave the details open, the agent improvises based on the existing scene and interaction context,} filling in spatial layouts, storylines, or NPC dialogue. For example, a player might open a door with no predefined content behind it. The agent can imagine ``a storage room with a locked iron chest in the corner'' and submit it to the engine to be registered in the world. From that point on, the engine maintains the chest's state and determines whether it can be opened.

\begin{figure*}[t]
  \centering
  \includegraphics[width=\linewidth]{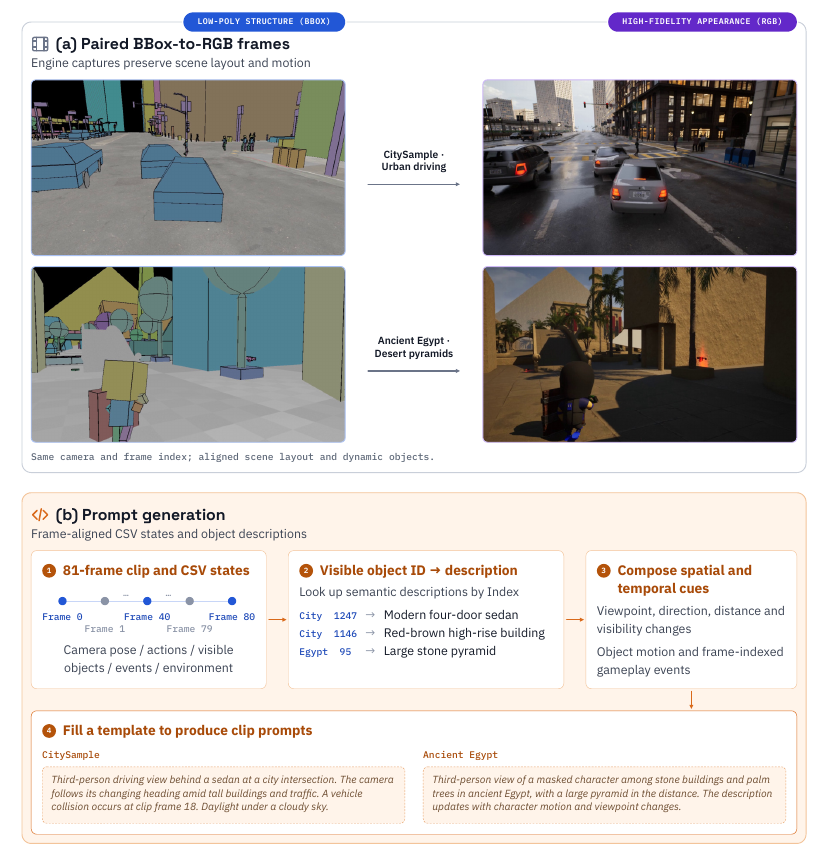}
  \caption{\textbf{UE5 paired training data.} (a) Each clip is captured twice in Unreal Engine 5 under the same camera and frame index: once as a low-poly structure render (BBox) that preserves scene layout and motion, and once as the high-fidelity RGB frame. (b) For every 81-frame clip, frame-aligned CSV records of camera pose, actions, visible objects, events, and environment are turned into a prompt: visible object IDs are looked up in a description table, spatial and temporal cues are composed, and a template produces the clip prompt.}
  \label{fig:data}
\end{figure*}

\subsection{Video Generation Model: The World's ``Painter''}
\label{sec:roadmap_video}

The video model transforms structural renderings into rich, detailed visuals. While preserving scene structure and motion constraints, it generates materials, lighting, atmosphere, and fine visual details, reducing the need to create high-fidelity 3D assets individually.

Conditioned on current control inputs and past frames, the model continuously generates video through causal autoregression. Few-step distillation and inference optimizations tailored to real-time re-rendering reduce the delay between player actions and visual feedback.

The real-time loop can be summarized as:
\begin{quote}
\small
\noindent
\textbf{Player input}
$\rightarrow$ \textbf{Engine updates world state}
$\rightarrow$ \textbf{Structural renderings and control signals}
$\rightarrow$ \textbf{Real-time video rendering}
$\rightarrow$ \textbf{Visual feedback to the player}.
\end{quote}

\paragraph{From Controllable Videos to Executable Worlds.}
The key is to separate how the game runs from how it looks: {code defines the rules, the engine maintains state, and the video model generates the visuals.} The engine only needs to run a lightweight simulation, leaving video generation as the main real-time performance challenge---one that can be mitigated through few-step distillation and causal autoregressive inference. Multiple players can share the same world state, each seeing it from their own perspective. As AI coding agents become more capable, the barrier to building lightweight games continues to fall, making this approach increasingly practical. Through this division of responsibilities, Matrix-Game V4 aims to help more people turn their ideas into games that can be played, modified, and continually refined---moving game world models from \textbf{``controllable videos'' to ``executable worlds.''}

\section{Data}
\label{sec:data}

\subsection{Paired Low-Poly and High-Fidelity Capture}
\label{sec:data_pairs}

We build a cross-domain dataset of aligned low-poly and high-fidelity videos in Unreal Engine 5 (UE5)~\cite{unrealengine} for world modeling and video generation. It covers four representative scene types: the CitySample megacity~\cite{epic2022citysample}, an open desert, a complex castle, and high-dynamic air combat. Table~\ref{tab:data} lists the training data of each scene. Every capture is recorded at $1280\times720$ and contains three aligned videos (the UE5 RGB render, the low-poly structure render, and a depth render of the proxy geometry) plus a per-frame metadata record. Shooter and air-combat clips last about one minute. For the castle, desert, and city scenes we hold out three complete capture sessions, and for air combat we hold out whole capture batches; no held-out session contributes any training clip.

\begin{table}[t]
  \centering
  \caption{\textbf{UE5 paired training data.} Each clip is a pair of pixel-aligned low-poly and high-fidelity videos at $1280\times720$. Held-out clips come from capture sessions (for air combat, whole multi-machine capture batches) that contribute no training clip.}
  \label{tab:data}
  \small
  \setlength{\tabcolsep}{5pt}
  \begin{tabular}{@{}llrrrr@{}}
    \toprule
    Scene & Game & Train clips & Hours & FPS & Held-out \\
    \midrule
    Medieval castle & Third-person shooter & 17,848 & 290 & 24 & 3 sessions, 683 clips \\
    Ancient Egypt desert & Third-person shooter & 40,296 & 643 & 24 & 3 sessions, 617 clips \\
    Desert canyon air combat & Flight combat & 28,720 & 467 & 24 & whole capture batches \\
    CitySample megacity & Driving & 27,470 & 312 & 17 & 3 sessions \\
    \midrule
    Total & & 114,334 & 1,712 & & \\
    \bottomrule
  \end{tabular}
\end{table}

The capture pipeline abstracts each high-complexity scene into a lightweight geometric representation. Camera trajectories and timestamps are strictly synchronized between the two renders, so the low-poly structure frame and the high-fidelity frame of the same index are aligned pixel by pixel (Figure~\ref{fig:data}a). The structure representation keeps the spatial topology, the rigid poses, and the motion trajectories of the scene. For buildings, complex terrain, vegetation, and dynamic interactive objects, adaptive geometric fitting and view-frustum culling remove material reflections, high-frequency textures, and complex lighting, so the structure prior describes only the essential spatial geometry. In the resulting proxy the sky is black and the ground is a gray checkerboard; buildings and walls become flat-colored boxes and prisms, trees become a cylinder with a sphere, characters become figures made of blocks, and every primitive is drawn with a black outline and no texture, reflection, or shadow (left part of each split frame in Figure~\ref{fig:teaser}). In parallel, the system records object instance IDs, visibility states, and causal interaction events, forming a semantic metadata stream for every clip. The data defines a conditional mapping from lightweight geometric topology to high-fidelity visuals, and gives deterministic supervision for spatiotemporal consistency and interaction stability in long rollouts.

\subsection{Dynamic Captions}
\label{sec:data_caption}

\definecolor{capevt}{HTML}{C85A10}
\definecolor{cappos}{HTML}{2257D6}
\newcommand{\cevt}[1]{\textcolor{capevt}{#1}}
\newcommand{\cpos}[1]{\textcolor{cappos}{#1}}
\providecommand{\figsub}[1]{{\microtypesetup{expansion=false}\sffamily\fontsize{7}{8.4}\selectfont #1}}  

\begin{figure*}[t]
  \centering
  \includegraphics[width=\linewidth]{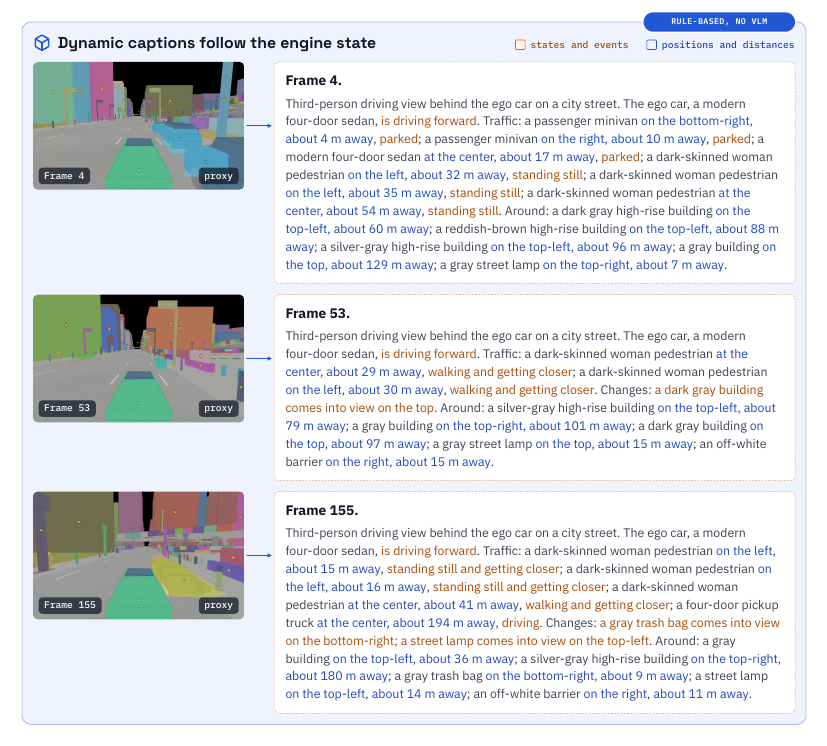}
  \caption{\textbf{Dynamic captions follow the engine state.} Top: low-poly proxy frames of one CitySample driving clip; numbered markers are the visible objects that the engine reports for this frame, each linked by its instance ID to a semantic description. Below: the caption that the rule-based template produces for the same frame, without any vision-language model. Orange text comes from object states and events, and blue text from object positions and distances. As the car moves, parked vehicles leave the description, pedestrians change from standing to approaching, and newly visible objects are announced under ``Changes''.}
  \label{fig:dyncap}
\end{figure*}

In an interactive game the picture changes with every player action. Turning the camera changes which objects are visible, and moving or triggering events changes the state of objects. A fixed text description cannot follow these changes, so Matrix-Game 4.0 uses dynamic captions: the description given to the video model is updated with the current state of the game world.

Each object in the scene has an ID that is linked to a semantic description. The system reads the currently visible objects and their real-time states from the game engine and fills them into a prompt template, producing a description that matches the current frame (Figure~\ref{fig:data}b). When the player changes the view, enters a new area, or triggers an event, the description is reorganized, so the video model always receives the latest semantic condition (Figure~\ref{fig:dyncap}). The dynamic caption and the low-poly proxy complement each other. The proxy supplies structure: position, layout, and motion. The caption supplies semantics: which objects are in the frame and what state each is in. Together they constrain generation, so the frames follow the scene structure produced by the engine and also follow changes in events and object states.

\section{Rendering Model}
\label{sec:model}

\begin{figure*}[t]
  \centering
  \includegraphics[width=\linewidth]{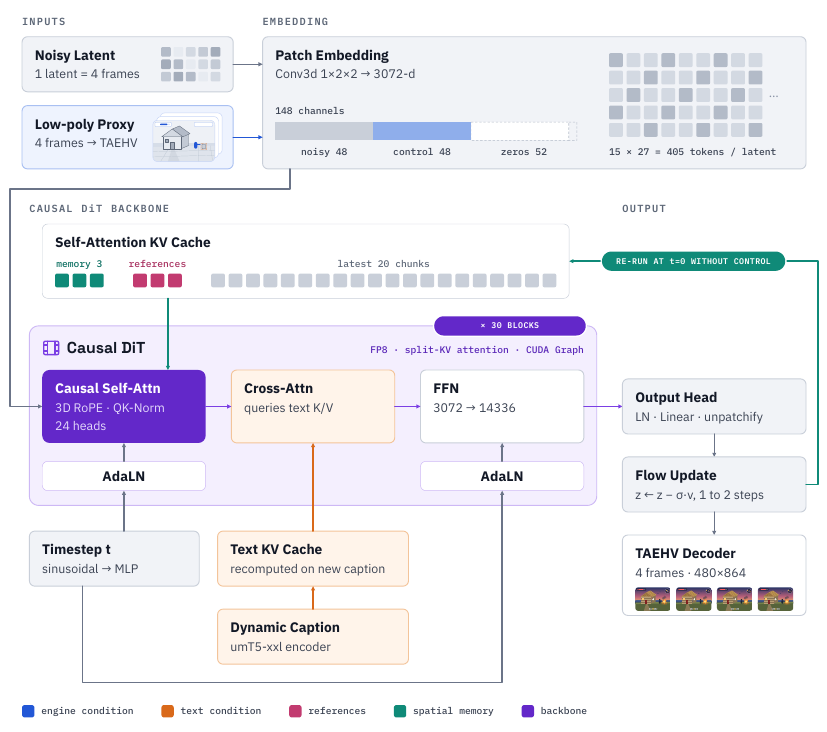}
  \caption{\textbf{Architecture of the rendering model.} The noisy latent of one chunk (one latent frame, four video frames) and the low-poly proxy encoded by TAEHV are concatenated into 148 channels and patchified into 405 tokens per latent frame at $480\times864$. Thirty causal DiT blocks apply self-attention over a KV cache that holds spatial-memory frames, reference frames, and the latest 20 chunks, cross-attention to the cached text keys and values of the dynamic caption, and a feed-forward network, each modulated by the timestep through AdaLN. The output head predicts a velocity, one or two flow updates produce the clean latent, and TAEHV decodes four frames. The finished chunk is then re-run at $t=0$ without the control input to write its keys and values into the cache.}
  \label{fig:arch}
\end{figure*}

\subsection{Architecture}
\label{sec:model_arch}

The rendering model starts from Wan2.2 5B~\cite{wan2025wan,wan2025wan22} in its Fun Control variant~\cite{alibabapai2025wan22fun} and is turned into a video-to-video (V2V) model. Four inputs condition each chunk, and each has its own job:
\begin{itemize}[leftmargin=*,itemsep=1pt,topsep=2pt]
  \item The \textbf{low-poly proxy} rendered by the engine is concatenated with the latent video being generated along the channel dimension. It tells the model what to draw and where.
  \item The \textbf{dynamic caption} (Section~\ref{sec:data_caption}) describes the scene and its current events and supplies the details.
  \item \textbf{Spatial-memory reference frames} keep the scene consistent when the player returns to a place.
  \item The \textbf{first frame} connects the generated video to what is already on screen.
\end{itemize}

Figure~\ref{fig:arch} shows the network, whose DiT has 5B parameters. The input of the patch embedding has 148 channels: 48 for the noisy latent, 48 for the control latent of the proxy, and 52 zero channels. A $1\times2\times2$ 3D convolution maps it to a width of 3072, which gives $15\times27=405$ tokens per latent frame at $480\times864$. The backbone has 30 DiT blocks. Each block applies causal self-attention, cross-attention to the text keys and values, and a feed-forward network with a hidden width of 14336, and the timestep modulates the block through AdaLN. Self-attention has 24 heads of width 128 with QK-Norm, and 3D RoPE assigns 44, 42, and 42 of the 128 dimensions to time, height, and width. The text encoder is umT5-xxl~\cite{chung2023unimax}. The key and value tensors of the caption are cached and recomputed only when the caption changes. The output head (LayerNorm, linear projection, unpatchify) predicts a velocity, and the latent is updated by $z \leftarrow z - \sigma\, v$ for one or two steps. TAEHV~\cite{boerbohan2025taehv} decodes the four frames of the chunk.

The self-attention KV cache has three parts: the keys and values of three spatial-memory frames, those of the reference frames, and those of the latest 20 chunks. After a chunk is generated, the model runs once more on the finished latent at $t=0$ without the control input and writes the resulting keys and values into the cache, so later chunks attend to the clean history.

\subsection{Supervised Fine-Tuning}
\label{sec:model_sft}
We train one model per game. Starting from the Wan2.2-Fun-5B Control weights, the model is fine-tuned as a bidirectional video-to-video model with the flow-matching objective of Wan, on 81-frame windows of the paired data at $704\times1280$. Each window is conditioned on its low-poly proxy, its caption, the first frame, and up to three context frames that serve as spatial memory. The context frames are taken from at least 40 frames before the window: candidates whose view overlaps the window are kept, and the closest ones by camera pose are selected. They enter the sequence as clean latents with timestep 0 and are excluded from the loss. To support classifier-free guidance and generation without a given first frame, the first frame is dropped with probability 0.3, the prompt with probability 0.2, and the proxy control with probability 0.2. Training uses 64 GPUs with a global batch of 64, AdamW with a learning rate of $10^{-5}$, and 4,000 steps.

\subsection{Causal Few-Step Distillation}
\label{sec:model_distill}
Following Causal Forcing++~\cite{zhao2026causalforcingplusplus}, we convert the bidirectional model into a causal autoregressive generator that denoises one chunk (one latent frame) at a time with one or two steps. The three stages below run at $480\times864$ on 81-frame windows with 64 GPUs and a global batch of 64.

\paragraph{Stage 1: teacher-forced autoregressive diffusion.}
Starting from the fine-tuned model, each latent frame receives its own noise level and attends only to the clean ground-truth frames before it, the first frame, and the context frames. The model is trained with the same flow-matching loss for 20,000 steps at a learning rate of $2\times10^{-6}$. The result is a causal teacher that samples each chunk with a multi-step ODE solver.

\paragraph{Stage 2: causal consistency distillation.}
A frozen copy of the causal teacher, an online student, and an EMA student start from the same weights. On a grid of 48 noise levels, the teacher takes one Euler step with classifier-free guidance of 3 from a noisy chunk to the next lower level; the online student predicts the clean chunk from the higher level and is trained to match the EMA student's prediction from the lower level. After 5,000 steps with an EMA decay of 0.99, the EMA student generates a chunk in four steps.

\paragraph{Stage 3: distribution matching.}
The student then learns from its own autoregressive rollouts with asymmetric DMD~\cite{yin2024one,yin2024improved,yin2025slow}. The real score is the frozen bidirectional model with classifier-free guidance, $v_{\text{c}} + 3\,(v_{\text{c}} - v_{\text{u}})$; the fake score is initialized from the same weights and trained online. The fake score is updated at every step and the student every five steps, with learning rates of $4\times10^{-7}$ and $2\times10^{-6}$. The student denoises each chunk in two steps and the first generated latent frame in four. Our latest recipe adds three changes that match the deployed inference setting. The student rolls out 21, 41, 61, or 121 latent frames with a rolling cache of the first frame and the latest 20 latent frames, and the two scores evaluate the last 21. At the start of every 20-frame segment, up to three spatial-memory frames are retrieved by camera pose and their keys and values join the cache. Because the deployed system encodes the proxy with TAEHV, the student receives TAEHV-encoded control on 80\% of the steps and Wan-VAE-encoded control on the rest, while both scores always see the Wan-VAE encoding. This stage runs for 1,000 steps from the end of Stage 2.

\section{Real-Time Inference}
\label{sec:inference}

\begin{figure}[t]
    \centering
    \includegraphics[width=\linewidth]{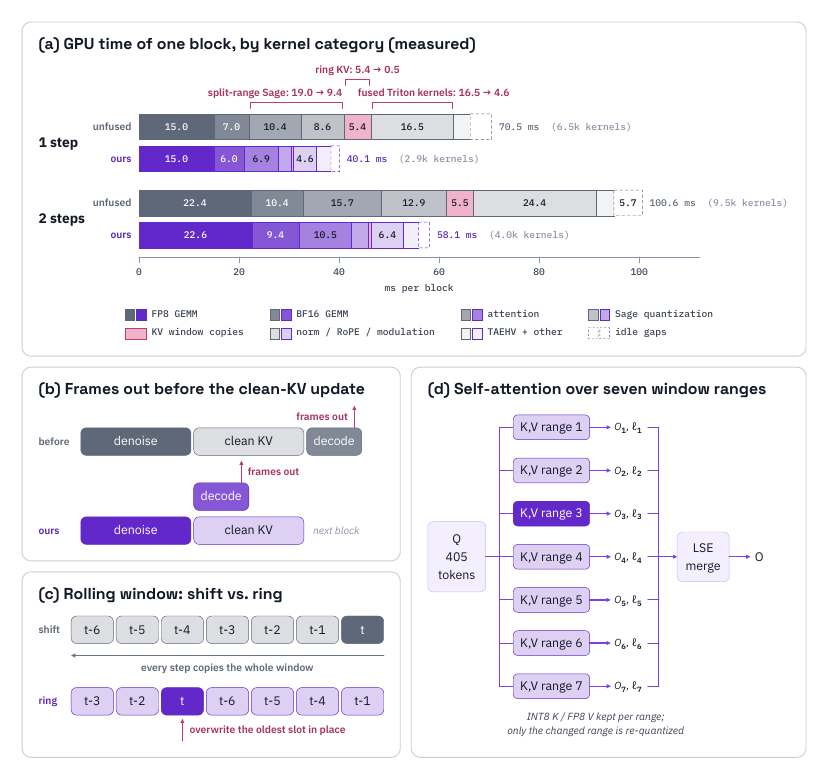}
    \caption{Overview of the inference optimizations. (a) Per-block GPU time on an RTX 5090,
    broken down by kernel category. (b) Decoding is moved ahead of the clean-KV update, so frames are output
    earlier. (c) Instead of shifting the whole KV window at every step, the ring cache writes the
    new frame into the oldest slot. (d) Self-attention is split into seven window ranges whose
    quantized keys and values are cached; the partial outputs are combined by log-sum-exp.}
    \label{fig:inference}
\end{figure}

\begin{table}[t]
  \centering
  \caption{\textbf{Inference speed and latency, applied cumulatively.} Each row adds one optimization to all rows above it. FPS is the steady-state generation rate of the model; latency is the time from a key press to the first generated frame that reflects it, with the game running at 60 FPS. Speedup is the FPS ratio to the previous row. Setting: $480\times864$, chunk size of one latent frame (four frames), KV window of 21 latent frames, one NVIDIA RTX 5090.}
  \label{tab:speed}
  \small
  \setlength{\tabcolsep}{6pt}
  \begin{tabular}{@{}lrrr@{}}
    \toprule
    Configuration & FPS $\uparrow$ & Latency (ms) $\downarrow$ & Speedup \\
    \midrule
    Causal Wan2.2 DiT, 4 steps, full Wan VAE & 4.5 & 1110 & -- \\
    + tiny VAE (TAEHV) & 6.5 & 867 & 1.4$\times$ \\
    + RoPE kernel, cached text KV & 15.4 & 551 & 2.4$\times$ \\
    + 8-bit weights + CUDA Graph & 24.2 & 466 & 1.6$\times$ \\
    + 1-step distillation & 51.2 & 376 & 2.1$\times$ \\
    + asynchronous I/O pipeline & 57.7 & 367 & 1.1$\times$ \\
    + fused Triton kernels & 65.8 & 121 & 1.1$\times$ \\
    + ring KV cache & 73.7 & 117 & 1.1$\times$ \\
    + split-KV FlashAttention & \textbf{83.4} & \textbf{112} & 1.1$\times$ \\
    \midrule
    Overall & 19$\times$ faster & 9.9$\times$ lower & \\
    \bottomrule
  \end{tabular}
\end{table}

The rendering model generates one chunk at a time, and a chunk is one latent frame, \ie four video frames. Short chunks keep the response to an action short, and they also make model calls frequent: with a single denoising step, each call does little arithmetic, and the GPU spends a large share of its time waiting for the CPU to launch kernels. We remove these costs in the order listed in Table~\ref{tab:speed}.

\paragraph{Launch overhead.}
The computation of the DiT backbone is recorded once as a CUDA Graph~\cite{nvidia2024cudagraphs} and replayed in a single launch at run time. Linear layers use 8-bit (FP8) weights~\cite{micikevicius2022fp8}.

\paragraph{Fused kernels.}
We write fused GPU kernels in Triton~\cite{tillet2019triton} for RoPE, normalization, and FP8 quantization. The query, key, and value projections of a layer share one quantization of their input. The key and value tensors of the text condition are cached and recomputed only when the caption changes.

\paragraph{KV cache and attention.}
The key/value cache of the video history is a ring buffer: when the window slides, the newest latent frame overwrites the oldest slot in place and no history is moved. Self-attention splits the KV window into parts that are computed in parallel and merged~\cite{dao2023flashdecoding}.

\paragraph{Decoding and I/O.}
The full Wan VAE decoder is replaced by the lightweight TAEHV decoder~\cite{boerbohan2025taehv}, and model input and output run in an asynchronous pipeline.

\paragraph{Result.}
Together these changes raise the steady-state rate from 4.5 to 83.4 FPS (19$\times$) and cut the key-to-frame latency from 1110 ms to 112 ms (9.9$\times$). Measuring model inference alone on one RTX 5090 at 480p, the steady-state generation rate is about 83 FPS with one denoising step and about 57 FPS with two, and the peak memory is about 18.2 GiB in both cases. With one step, the model denoises one four-frame chunk in 48 ms. The text encoder, the DiT, and the VAE all reside on the same card.

We further optimize and evaluate the renderer on 16 GiB consumer GPUs. At
$480\times864$ with one denoising step and fixed prompts,
steady-state generation throughput reaches approximately 67 FPS
on an RTX 5080 and 58 FPS on an RTX 5070 Ti, with GPU memory usage of approximately 12.8 GiB and 10.9 GiB, respectively.
\section{Evaluation}
\label{sec:evaluation}

\subsection{Setup}
\label{sec:exp_setup}
We train one rendering model per game on the data in Table~\ref{tab:data}. The results in this section use the distilled causal models of the castle shooter, the desert shooter, and city driving, each generating two denoising steps per chunk. A held-out rollout starts from the ground-truth first frame and runs autoregressively for 961 frames (40 seconds at 24 FPS) at $480\times864$, conditioned on the low-poly proxy and the caption of a held-out clip. Speed and latency are measured on a single NVIDIA RTX 5090 (Section~\ref{sec:inference}).

\subsection{Comparison with Recent Interactive Systems}
\label{sec:exp_compare}

\begin{table}[t]
  \centering
  \caption{\textbf{Latency, hardware, and openness of recent real-time interactive generation systems.} Latency is the end-to-end latency, and resolution is the output resolution at which each system reports its timings. For the systems that generate in chunks, the latency is two chunk generation times. Code World Model (streaming) renders a 17-frame chunk in 5.1 s on one H200. For StreamRender-H3, the chunk time of 258 ms is the sum of the three warm-renderer stage times reported on its project page: 16.67 ms for reference encoding, 217.96 ms for the two-step render, and 23.8 ms for causal video decoding, the last measured at 720p on one GB200. AgentGarten renders a 16-frame chunk in 439 ms on one H100. For Magpie, the latency is the end-to-end time reported by its authors: 0.83 s for the engine to prepare a 20-frame chunk, about 0.1 s for transmission and encoding, and 0.62 s for rendering. Matrix-Game 4.0 latency is measured from a key press to the first frame that reflects it, with the game running at 60 FPS. Open release is checked against the GitHub and Hugging Face pages of each project on October 19, 2026.}
  \label{tab:compare}
  \small
  \setlength{\tabcolsep}{4pt}
  \begin{tabular}{@{}lll>{\raggedright\arraybackslash}p{3.8cm}>{\raggedright\arraybackslash}p{3.4cm}@{}}
    \toprule
    System & Latency & Resolution & Compute & Open release \\
    \midrule
    \begin{tabular}[t]{@{}l@{}}Code World Model\\(streaming)~\cite{chen2026codeworldmodel,liao2026streamingcwm}\end{tabular} & $\approx$10.2 s & 1344$\times$768 & 1$\times$ H200 141 GB; peak memory 108--123 GB, does not fit on 80 GB cards & Code and LoRA weights; 40 example proxies, proxy generator not released \\
    StreamRender-H3~\cite{lu2026streamrenderh3} & $\approx$517 ms & 1344$\times$768 & 8$\times$ GB200 with context parallelism & Code, weights, and the racing engine with its track and car assets \\
    AgentGarten~\cite{chi2026agentgarten} & $\approx$878 ms & 832$\times$480 & 1$\times$ H100 80 GB (data-center GPU) & Renderer code and weights; real-time engine and code worlds not released \\
    Magpie~\cite{zhan2026magpie} & $\approx$1.55 s & 1280$\times$768 & 1$\times$ H100 80 GB (data-center GPU), peak memory about 34 GB & Lite dataset of render and white-box videos; code and weights not released \\
    \midrule
    \textbf{Matrix-Game 4.0} & \textbf{112 ms} & 864$\times$480 & \textbf{1$\times$ RTX 5090}, a single consumer GPU with 32 GB memory & Code and weights, released at launch \\
    \bottomrule
  \end{tabular}
\end{table}

Table~\ref{tab:compare} compares Matrix-Game 4.0 with four recent systems that generate interactive video from structured or code-defined worlds. Matrix-Game 4.0 has the lowest end-to-end latency, about one fifth of that of the next system, and it is the only one that runs on a single consumer GPU.

\subsection{Qualitative Results}
\label{sec:exp_qual}

\providecommand{\figsub}[1]{{\microtypesetup{expansion=false}\sffamily\fontsize{7}{8.4}\selectfont #1}}  

\begin{figure*}[t]
  \centering
  \includegraphics[width=\linewidth]{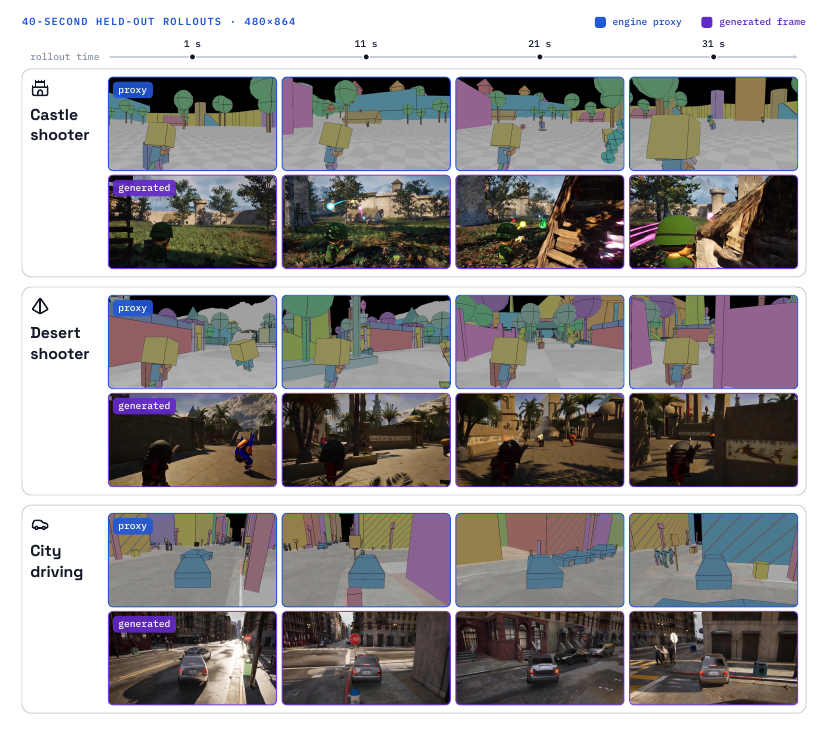}
  \caption{\textbf{Re-rendering over 40-second held-out rollouts.} For each game, the upper row is the low-poly proxy from the engine and the lower row is the frame generated by the distilled causal model at the same time step. The generated frames follow the proxy layout, camera, and object motion throughout the rollout, while the model fills in materials, vegetation, lighting, smoke, and muzzle effects.}
  \label{fig:qual}
\end{figure*}

Figure~\ref{fig:qual} shows three held-out rollouts at four time steps. The generated frames keep the layout of the proxy: buildings, walls, trees, and vehicles appear where the proxy places them, and the camera follows the same path. The model adds what the proxy omits. Flat-colored blocks become stone walls, thatched houses, palm trees, sandstone buildings, and city facades; the box-shaped player becomes a textured character; and combat produces smoke, sparks, and projectile trails. The look of each scene stays stable over the 40 seconds, and the character keeps its appearance from the first chunk to the last.

\subsection{Gameplay Customization with the Coding Agent}
\label{sec:exp_play}

\begin{figure*}[t]
  \centering
  \includegraphics[width=\linewidth]{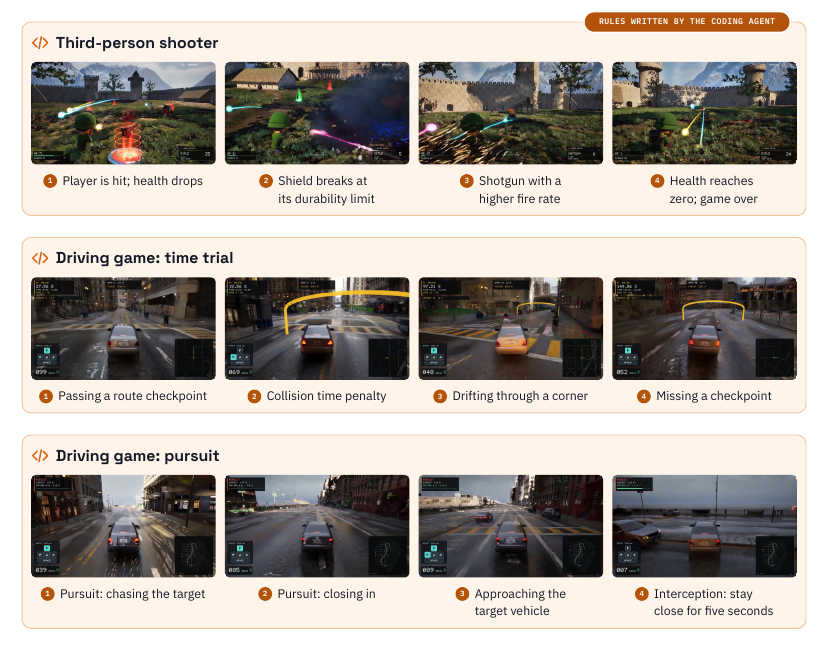}
  \caption{\textbf{Game modes written with the coding agent, rendered in real time.} Top: in the third-person shooter, the agent adds health and ammunition control, shield durability, a faster shotgun, a game-over condition, and simple sound effects. Middle: in the driving game, the agent implements a time trial with route checkpoints and collision penalties. Bottom: the same driving game turned into a pursuit mode, in which the player must intercept a target vehicle and stay close to it for five seconds. The engine evaluates every rule; the video model renders each frame from the engine output.}
  \label{fig:play}
\end{figure*}

With the coding agent, a player can customize a game in natural language on top of the existing game code: reshape the terrain, place scene objects, change NPC behavior and interaction logic, or design an entirely new mode. A shooter where everyone fights alone can become a team battle, and a classic time trial can become a cops-and-robbers chase. Figure~\ref{fig:play} shows three modes built this way. The shooter gains health and ammunition control, a shield with limited durability, a shotgun with a higher fire rate, a game-over condition at zero health, and simple sound effects. The driving game gains a time trial with checkpoints, collision penalties, and drifting, and a pursuit mode in which the player intercepts a target vehicle. All of these rules run in the engine, and the video model renders whatever the engine outputs. We encourage creators to adapt and extend the released game code to explore further gameplay.

\section{Related Work}
\label{sec:related}

\subsection{Interactive Video World Models}
Interactive world models generate the next observations of an environment from user actions~\cite{chen2026codeworldmodel,liao2026streamingcwm,huang2026pwm,liang2026codevideomodel,zhan2026magpie,lin2026alayarendererflash,meshy2026mora,chi2026agentgarten,lu2026streamrenderh3,fang2026solarrenderer,huang2026edenlive,yu2025egosim,qian2026matrix,yu2026mosaicmem}. Genie learns a latent action space and an autoregressive dynamics model from unlabeled Internet videos~\cite{bruce2024genie}; Genie~2 and Genie~3 generate playable 3D environments, and Genie~3 does so in real time at 720p and 24 FPS~\cite{parkerholder2024genie2,genie3}. GameNGen runs DOOM as a diffusion model at 20 FPS on a single TPU~\cite{valevski2024diffusion}, DIAMOND trains agents inside a diffusion world model~\cite{alonso2024diffusion}, and Oasis generates a Minecraft-like game in real time~\cite{oasis2024}. Later systems add richer controls and longer horizons: Hunyuan-GameCraft maps keyboard and mouse input to a shared camera representation~\cite{gamecraft,gamecraft2}, Yan targets real-time 1080p simulation and text-driven editing~\cite{ye2025yan}, WorldPlay and LingBot-World stream minute-scale interactive video~\cite{sun2025worldplay,team2026advancing}, and Matrix-Game~2.0 and~3.0 combine few-step autoregressive diffusion with long-horizon memory for real-time 720p generation~\cite{he2025matrix,wang2026matrixgame3}. In these models the network carries both the world dynamics and the visuals. Matrix-Game 4.0 keeps the dynamics in game code and uses the network for the visuals.

\subsection{Code-Defined Worlds Rendered by Video Models}
A recent line of work assigns state and rules to executable code and rendering to a video model. In Code World Model, a coding agent writes code that maintains the world state, and a proxy video compiled from that state conditions a fine-tuned MiniMax-H3 renderer~\cite{chen2026codeworldmodel}. Streaming Code World Model turns this renderer into a causal streaming generator distilled to four steps with DMD, which generates each 17-frame chunk in 5.1~s on one H200~\cite{liao2026streamingcwm}. Programmable World Model has an agent translate instructions into programs of entity states and transition rules, executes them in a lightweight engine that also keeps off-screen entities, and renders the compiled 3D bounding boxes through a ControlNet attached to a pretrained video model~\cite{huang2026pwm}. Code Video Model writes Three.js programs that lay out scenes, object motion, and cameras, and conditions a frozen MiniMax-H3 on the resulting proxy video only in the early denoising steps, without additional training~\cite{liang2026codevideomodel}. StreamRender-H3 drives a two-step streaming renderer built on MiniMax-H3 with the semantic maps of a code game engine~\cite{lu2026streamrenderh3}, on top of the Sol inference engine~\cite{li2026sol}. AgentGarten couples simulators and game engines with a shared neural renderer and uses the resulting worlds to train agents~\cite{chi2026agentgarten}. CoDeR builds executable worlds in code through coordinated roles and renders them with video generation models~\cite{fang2026coder}, and Magpie separates a game engine that resolves player actions from a render server that generates visuals from engine frames~\cite{zhan2026magpie}. Mora combines a coding agent, 3D generation, and a video model to build puzzle, multiplayer, and physics games that are played in the cloud~\cite{meshy2026mora}, and EdenLive renders the reference videos produced by upstream game programs into detailed frames with a four-step block-autoregressive model~\cite{huang2026edenlive}. Matrix-Game 4.0 follows the same division of labor and is designed for interactive latency on one consumer GPU: it reaches 112 ms from key press to frame on a single RTX 5090 (Table~\ref{tab:compare}), and it adds a coding agent that edits the game rules and writes the prompts during play.

\subsection{Neural Rendering and Structure-Conditioned Video Generation}
Neural rendering combines learned generative models with the structure of the graphics pipeline~\cite{tewari2020state,tewari2022advances}. Enhancing Photorealism Enhancement translates rendered game frames toward photographs using G-buffers from the renderer~\cite{richter2023enhancing}, GameGAN learns a game renderer with a memory module from frames and actions~\cite{kim2020gamegan}, and DLSS~5 adds photoreal lighting and materials to game frames in real time~\cite{nvidia2026dlss5}. vid2vid and World-Consistent vid2vid convert semantic maps into temporally coherent video, the latter keeping revisited regions consistent through projected 3D guidance~\cite{wang2018vid2vid,mallya2020world}. With diffusion models, ControlNet and VideoComposer add spatial and temporal controls to pretrained generators~\cite{zhang2023controlnet,wang2023videocomposer}; Cosmos-Transfer1 turns segmentation, depth, and edge maps into realistic video~\cite{nvidia2025cosmostransfer1}; and DiffusionRenderer and FrameDiffuser synthesize video from G-buffers~\cite{liang2025diffusionrenderer,beisswenger2025framediffuser}. AlayaRenderer-Flash distills a G-buffer-conditioned renderer into a four-step autoregressive model that runs at 31.5~FPS at $832\times448$ on one H200~\cite{lin2026alayarendererflash}, and SolarRenderer follows arbitrary structure inputs, including skeletons, sketches, and point clouds, through the reference pathway of MiniMax-H3 and renders them in four steps~\cite{fang2026solarrenderer}. Our rendering model starts from the Wan2.2 Fun Control weights~\cite{alibabapai2025wan22fun,aigcapps2026videoxfun}, conditions on the low-poly proxy video rendered by the game engine, and is trained on pixel-aligned UE5 pairs.

\subsection{Real-Time Autoregressive Video Generation}
Streaming generation produces video chunk by chunk. Diffusion Forcing and Rolling Diffusion assign different noise levels across frames~\cite{chen2024diffusion,ruhe2024rolling}, and StreamDiT trains on a moving frame buffer for real-time text-to-video~\cite{kodaira2026streamdit}. Distribution matching distillation (DMD) turns a multi-step diffusion model into a one-step or few-step generator~\cite{yin2024one,yin2024improved}. CausVid applies DMD to convert a bidirectional video model into a causal one~\cite{yin2025slow}; Self Forcing and Self-Forcing++ train on the student's own KV-cached rollouts~\cite{huang2025self,cui2025self}; Causal Forcing initializes the student from an autoregressive teacher~\cite{zhu2026causal}; and Causal Forcing++ uses causal consistency distillation to reach frame-wise generation with one or two steps~\cite{zhao2026causalforcingplusplus}. We follow Causal Forcing++ to distill our bidirectional rendering model. At the systems level, CUDA Graphs~\cite{nvidia2024cudagraphs}, FlashAttention and split-KV decoding~\cite{dao2022flashattention,dao2024flashattention2,dao2023flashdecoding}, FP8 arithmetic~\cite{micikevicius2022fp8}, Triton kernels~\cite{tillet2019triton}, and the TAEHV decoder~\cite{boerbohan2025taehv} supply the components of our inference stack.

\section{Conclusion and Future Work}
\label{sec:conclusion}



Matrix-Game~4.0 separates game execution from visual generation
through an executable game harness. Game code defines the rules,
a lightweight engine maintains persistent world state, and a video
model generates the visuals in real time. A coding agent connects
user ideas to executable game code, enabling scenes, interactions,
and gameplay to be created and iteratively refined through
natural-language instructions and interactive play.

On a single NVIDIA RTX~5090, the renderer achieves 83~FPS
steady-state generation throughput at 480p with one denoising step.
The integrated system achieves a measured key-press-to-display
latency of 112~ms in our evaluated configuration. The pipeline
spans paired UE5 gameplay data, dynamic semantic conditioning,
causal adaptation, few-step distillation, and optimized inference,
providing a reusable foundation for adapting this approach to
additional games. For researchers, Matrix-Game~4.0 offers a
technical basis for investigating this emerging paradigm.
For game developers, it provides a concrete preview of a
game-development workflow that combines natural-language
prototyping, engine-enforced mechanics, and neural rendering.

In the near term, we will focus on reducing inference cost and
hardware requirements, with the goal of releasing a technical
preview on Steam. Further directions include improving visual
fidelity, incorporating a refinement stage for higher-resolution
output, exploring multiplayer rendering with consistent views
across players, enabling more precise control over generated
visuals, and extending the system to a broader range of game
genres and scenes.

\clearpage
\appendix

\FloatBarrier

{
\small
\bibliography{neurips_2025}
\bibliographystyle{unsrt}
}

\end{document}